%% file: main.tex
\documentclass[runningheads]{llncs}
\usepackage{graphicx}
\usepackage{cite}
\usepackage{rotating}
\usepackage{booktabs}
\usepackage{hyperref}
\usepackage{multirow}
\begin{document}

\title{Drone-based AI and 3D Reconstruction for Digital Twin Augmentation}

%
\author{Alex To\inst{1} \and
Maican Liu\inst{2} \and
Muhammad Hazeeq Bin Muhammad Hairul\inst{2} \and \\
Joseph G. Davis\inst{1} \and
Jeannie S.A. Lee\inst{3} \and
Henrik Hesse\inst{4} \and
Hoang D. Nguyen\inst{5}}
\authorrunning{To et al.}
%
\institute{School of Computer Science, University of Sydney, Australia\\
\email{duto3894@uni.sydney.edu.au,joseph.davis@sydney.edu.au} \and
Aviation Virtual Pte Ltd, Singapore\\
\email{liumaican@gmail.com,hazeeq\_hairul@hotmail.com} \and
Infocomm Technology, Singapore Institute of Technology, Singapore\\
\email{Jeannie.Lee@singaporetech.edu.sg} \and
James Watt School of Engineering, University of Glasgow, Singapore\\
\email{Henrik.Hesse@glasgow.ac.uk} \and
School of Computing Science, University of Glasgow, Singapore\\
\email{Harry.Nguyen@glasgow.ac.uk}
}
\maketitle              
\setlength{\emergencystretch}{3em}
\input{document}

%
%
%
\bibliographystyle{splncs04}
\bibliography{references}
\end{document}

%% file: document.tex
\keywords{Digital Twin \and 3D Reconstruction \and Artificial Intelligence (AI) \and Information Augmentation \and Unmanned Aerial Vehicle (UAV)}

\begin{abstract}
Digital Twin is an emerging technology at the forefront of Industry 4.0, with the ultimate goal of combining the physical space and the virtual space. To date, the Digital Twin concept has been applied in many engineering fields, providing useful insights in the areas of engineering design, manufacturing, automation, and construction industry. While the nexus of various technologies opens up new opportunities with Digital Twin, the technology requires a framework to integrate the different technologies, such as the Building Information Model used in the Building and Construction industry. In this work, an Information Fusion framework is proposed to seamlessly fuse heterogeneous components in a Digital Twin framework from the variety of technologies involved. This study aims to augment Digital Twin in buildings with the use of AI and 3D reconstruction empowered by unmanned aviation vehicles. We proposed a drone-based Digital Twin augmentation framework with reusable and customisable components. A proof of concept is also developed, and extensive evaluation is conducted for 3D reconstruction and applications of AI for defect detection. 
\end{abstract}

\section{Introduction}\label{introduction}

A Digital Twin is the virtual replication of a physical object. Through
modelling and real-time data communication, the Digital Twin simulates
the actual properties and behaviours of its physical counterpart in the
physical space, thus, enable learning, reasoning, and dynamically
re-calibrating for improved decision-making
\cite{glaessgen2012digital,grieves2014digital}. The tight and seamless
integration between the physical and virtual space in the Digital Twin
paradigm makes it one of the most promising enabling technologies for
the realization of smart manufacturing and Industry 4.0
\cite{tao2018digital}

To date, Digital Twin applications have seen success in various
industries and domains, including product design, production, prognostic
and health management, building and construction, and many others.
Recent advances in sensor technologies, big data, cloud computing,
social networks, Internet of Things (IoT), Computer-Aided-Design (CAD),
3D modelling, and Artificial Intelligence (AI) allow a massive amount of
data to be collected while enabling real-time communication for the
realization of the Digital Twin paradigm throughout the complete
product's life-cycle
\cite{tao2018digital,qi2019enabling,tao2018digitalb, fuller2020digital}.

In the Building and Construction industry context, physical objects are
buildings and structural components. To generate and capture their
virtual counterparts in the virtual space, Building Information Model
(BIM) is a common standard that encompasses a large amount of detail on
building dimensions and critical components. These components include
façade features, dimensions of staircases, slopes of walls, height of
railings, etc. The use of BIM provides high-quality preconstruction
project visualisation, improved scheduling, and better coordination and
issue management. The Digital Twin paradigm in this industry utilises
BIM as one of the core technologies to facilitate information
management, information sharing, and collaboration among stakeholders in
different domains over the building life cycle
\cite{adamu2015social,das2014bimcloud}.

In many cases, it is often desirable to obtain 3D models of the physical
buildings and landscapes that can be used for enrichment, visualization,
and advanced analytics of Digital Twin models
\cite{minos2018towards,henry2014rgb,spreitzer2020large,dryanovski2017large}.
Additionally, other sources of information can be useful for the Digital
Twin models, such as contextual information and geographical information
systems (GIS). However, there are some limitations with the current BIM
technologies that hinder the capability to integrate multiple sources of
data. For instance, BIM files are restricted in size, making it
difficult to add large artefacts. The BIM format is neither initially
designed for the integration of heterogeneous data sources nor capable
of capturing real-time updates.

The process to obtain 3D models of the physical buildings and landscapes
is also labour-intensive. When 3D reconstruction is done manually using
a hand-held device, the ability to capture an extensive model of the
building is often limited due to physical constraints such as the size
of tunnels or large constructions. These issues call for a new and more
scalable approach in 3D reconstruction using unmanned aerial vehicles
(UAV), optimal scanning methods, and advanced onboard processing
algorithms
\cite{mauriello2014towards,spreitzer2020large,shim2019development,Chan2021}.

On the other hand, real-time applications of AI and image analysis of
BIM is also underdeveloped. One application of imaging in building
maintenance for Digital Twin is defect detection, in which AI algorithms
are employed to recognize defect regions such as cracks automatically.
To develop such AI models, appropriate training data is required which,
however, is often found in 2D format, prompting suitable methods on
real-time transformation for AI applicability.

Motivated by the current limitations of building and construction
technologies, we aim to innovate BIM for Digital Twin in two broad
areas: 1) Develop an Information Fusion framework that extends BIM with
a metadata layer to support heterogeneous data integration; 2) Enhance
real-time synchronization between the physical space and virtual space
in BIM through improved 3D reconstruction methods and real-time
scanning.

To this aim, our approach is four-fold: First, we developed a proof of
concept Information Fusion framework to facilitate the integration of
multiple sources of information to produce useful data representations
for BIM applications. It utilises a distributed and fault-tolerant
database to store geometry objects (e.g., buildings and structural
components) and meta-information (e.g., defects and tagged items) to
provide maximal compatibility and highest/raw details; Second, we built
a drone-based 3D reconstruction solution for scalable data collection
and evaluate major scanning technologies including Light Detection and
Ranging (LiDAR) sensor, stereovision, and single-lens camera; Third, we
tested our real-time scanning capabilities by performing real-time 2D to
3D mapping from our camera feed at five frames per second. The mapping
computation is done on the drone using an onboard miniature computer;
Finally, we presented a defect detection use-case as an application of
AI in real-time image scanning.

The contributions of our work are as follows. We provided a
comprehensive review of Digital Twin technologies in conjunction with
AI. We demonstrated an end-to-end proof of concept of the use of BIM for
Digital Twin and information fusion. We conducted extensive experiments
for the evaluation of 3D reconstruction techniques. Finally, we
illustrated the feasibility of AI application in Digital Twin through
defect detection with deep learning use-case. Our work provides some
insights and theoretical and empirical implications for researchers as
well as practitioners in this emerging field.

\section{Background}\label{background}

\subsection{Digital Twin Technologies and
Applications}\label{digital-twin-technologies-and-applications}

The concept and model of the Digital Twin were publicly introduced in
2002 by Grieves in his presentation as the conceptual model underlying
Product Lifecycle Management \cite{grieves2002conceptual}. Although the
term was not coined at that time, all the Digital Twin's basic elements
were described: physical space, virtual space, and the information flow
between them. The key enablers of Digital Twin: sensor technologies,
cloud computing, Big Data, IoT, and AI have since then experienced
growth at an unprecedented rate. Recently, the concept of Digital Twin
was formally defined by NASA as a multiphysics, multiscale,
probabilistic, ultra-fidelity simulation that enables real-time
replication of the state of the physical object in cyberspace based on
historical and real-time sensor data.

Tao et al. extended the model and proposed that Digital Twin modelling
should involve: physical modelling, virtual modelling, connection
modelling, data modelling, and service modelling \cite{tao2018digital}.
From a more structural and technological viewpoint, Digital Twin
consists of sensor and measurement technologies, IoT, Big Data and AI
\cite{kaur2020convergence,kusiak2017smart}.

The applications of Digital Twins span various domains from
manufacturing, aerospace to cyber-physical systems, architecture,
construction, and engineering.

\subsubsection{Digital Twin in
Manufacturing}\label{digital-twin-in-manufacturing}

Applications of Digital Twin are prominent in smart manufacturing. Due
to ever-increasing product requirements and rapidly changing markets,
there has been a growing interest in shifting problem identification and
solving to early stages of product development lifecycle (also known as
``front-loading'') \cite{thomke2000effect}. The Digital Twin paradigm
fits perfectly because virtual replications of physical products allow
early feedback, design changes, quality, and functional testing without
entering the production phase.

Tao et al. suggested that a Digital Twin-driven product design process
can be divided into conceptual design, detailed design, and virtual
verification \cite{tao2019digital}. Throughout the process, various
kinds of data such as customer satisfaction, product sales, 3D model,
product functions, and configuration, sensor updates can be integrated
to mirror the life of the physical product to its corresponding digital
twin. With real-time closed-loop feedback between the physical and the
virtual spaces, designers are able to make quick decisions on product
design adjustment, quality control and improve the design efficiency by
avoiding tedious verification and testing.

During production, simulation of production systems, the convergence of
the physical and virtual manufacturing world leads to smart operations
in the manufacturing process, including smart interconnection, smart
interaction, smart control, and management. For example, Tao et al.
proposed a shop-floor paradigm consists of four components physical
shop-floor, virtual shop-floor, shop-floor service system driven by
shop-floor digital twin data, enabled by IoT, big data, and artificial
intelligence \cite{tao2017digital}. Modeling of machinery, manufacturing
steps, and equipment also help in precise process simulation, control,
and analysis, eventually leading to improvement of the production
process \cite{botkina2018digital}. A similar effort is observed in
\cite{minos2018towards} to evaluate different methods in automated 3D
reconstruction in SME factories. The authors explored the use of
low-cost stereo vision techniques with Simultaneous Localization and
Mapping (SLAM) to generate Digital Twin models of a physical factory
floor and machinery.

\subsubsection{Digital Twin in Building and
Construction}\label{digital-twin-in-building-and-construction}

Modelling physical buildings and landscapes with Digital Twin brought
valuable opportunities to the architecture, construction, and
engineering industry, such as improvements in urban planning, city
analytics, environmental analysis, building maintenance, defect
detection, and collaboration between stakeholders. An important concept
in this domain is BIM \cite{kensek2014building}, i.e.~a process
involving the generation and management of digital representations of
physical and functional characteristics of places.

Yan et al. proposed a method for the integration of 3D objects and
terrain in BIMs supporting the Digital Twin, which takes the accurate
representation of terrain and buildings into consideration
\cite {yan2019integration}. The authors discussed topological issues
that can occur when integrating 3D objects with terrain. The key to
solving this issue lies in obtaining the correct Terrain Intersection
Curve (TIC) and amending 3D objects and the terrain properly based on
it. Models developed by such methods are used for urban planning, city
analytics, or environmental analysis.

For preventive maintenance of prestressed concrete bridges, Shim et al.
proposed a new generation of the bridge maintenance system by using the
Digital Twin concept for reliable decision-making
\cite{shim2019development}. 3D models of bridges were built to utilise
information from the entire lifecycle of a project by continuously
exchanging and updating data from stakeholders

Digital Twin also finds application in recording and managing cultural
heritage sites. The work by \cite{dore2012integration} integrated a 3D
model into a 3D GIS and bridge the gap between parametric CAD modeling
and 3D GIS. The final model benefits from both systems to help document
and analyze cultural heritage sites.

From most construction projects, the presence of BIM is prominent due to
its wide range of benefits. BIM has received considerable attention from
researchers with works aiming to improve or extend its various aspects
for e.g social aspect \cite{adamu2015social}, elasticity and scalability
\cite{das2014bimcloud}, sustainability \cite{krygiel2008green}, safety
\cite{zhang2013building} and many others.

\subsubsection{Digital Twin in Smart
Nations}\label{digital-twin-in-smart-nations}

Gartner's Top 10 Strategic Technology Report for 2017 predicted that
Digital Twin is one of the top ten trending strategic technologies
\cite{Panetta2016}. Digital Twin since 2012 has entered rapid growth
stage considering the current momentum with applications in several
industries and across variety of domains.

NASA and U.S Air Force adopted Digital Twin to improve production of
future generations of vehicles to become lighter while being subjected
to high loads and more extreme service conditions. The paradigm shift
allowed the organisation to incorporate vehicle health management
system, historical data and fleet data to mirror the life of its flying
twin, thus, enabled unprecedented levels of safety and reliability
\cite{glaessgen2012digital}.

The world's 11th busiest airport, the second largest in the Netherlands,
Amsterdam Airport Schiphol built a digital asset twin of the airport
based on BIM. Known as the Common Data Environment (CDE), Schiphol's
Digital Twin solution integrates data from many sources: BIM data; GIS
data; and data collected in real-time on project changes and incidents
as well as financial information, documents, and project portfolios. The
information fusion capability of Digital Twin presents opportunities to
run simulations on potential operational failures throughout the entire
complex \cite{Baumann2019}.

Port of Rotterdam built a Digital Twin of the port and used IoT and
artificial intelligence to collect and analyse data to improve
operations. Digital Twin helps to better predict accurately what the
best time is to moor, depart and how much cargo needs to be unloaded.
Furthermore, real-time access to information enables better prediction
of visibility and water conditions \cite{Boyles2019}.

\subsection{Artificial Intelligence in Digital
Twin}\label{artificial-intelligence-in-digital-twin}

The rapid adoption of enabling technologies such as IoT, cloud
computing, and big data opens up endless opportunities for AI
applications in Digital Twin. As a multidisciplinary field, AI
encompasses Machine Learning, Data Mining, Computer Vision, Natural
Language Processing, Robotics, among many others. AI emerges as a
promising core service in Digital Twin to assist humans in decision
making by finding patterns, insights in big data, generation of
realistic virtual models through advanced computer vision, natural
language processing, robotics, etc.

Li et al. proposed a method that uses a concept of dynamic Bayesian
networks for Digital Twin to build a health monitoring model for the
diagnosis and prognosis of each individual aircraft
\cite{li2017dynamic}. For example, in diagnosis by tracking
time-dependent variables, the method could calibrate the
time-independent variables; in prognosis, the method helps predict crack
growth in the physical subject using particle filtering as the Bayesian
inference algorithm.

In production, \cite{alexopoulos2020digital} introduced a Digital
Twin-driven approach for developing Machine Learning models. The models
are trained for vision-based recognition of parts' orientation using the
simulation of Digital Twin models, which can help adaptively control the
production process. Additionally, the authors also proposed a method to
synthesize training datasets and automatic labelling via the simulation
tools chain, thus reducing users' involvement during model training.

Chao et al. \cite{fan2021disaster} described an insightful vision of
Digital Twin to enable the convergence of AI and Smart City for disaster
response and emergency management. In this vision, the authors listed
four components in Disaster City Digital Twin, i.e.~1) multi-data
sensing for data collection, 2) data integration and analytics, 3)
multi-actor game-theoretic decision making, 4) dynamic network analysis,
and elaborated the functions that AI can improve within each component.

Another interesting vision of Digital Twin in Model-Based Systems
Engineering is described in \cite{madni2019leveraging} in which the
realization of Digital Twin is progressively divided into four levels 1)
Pre-Digital Twin, 2) Digital Twin, 3) Adaptive Digital Twin and 4)
Intelligent Digital Twin. In the last two levels: Adaptive Digital Twin
and Intelligent Digital Twin, the authors emphasized the tight
integration of AI in engineering processes; for example, in level 3, an
adaptive user interface can be offered by using supervised machine
learning to learn the preferences and priorities of human operators in
different contexts, therefore, support real-time planning and decision
making during operators, maintenance and support; in level 4,
additionally unsupervised machine learning can help discern objects, and
patterns in the operational environment and reinforcement learning can
learn from continuous data stream from the environment.

Power networks are the backbone of power distribution, playing a central
economical and societal role by supplying reliable power to industry,
services, and consumers. To improve the efficiency of power networks,
researchers in the Energy industry have also been putting initial effort
into integrating Digital Twin, and AI for informed decision-making in
operation, support, and maintenance \cite{marot2020l2rpn}. In
particular, a virtual replication of the power network is developed.
Various time-series measurements from the physical power networks, such
as production values, loads, line thermal limits, power flows, etc., are
streamed back to the virtual models. Based on the digital models,
researchers exploit machine learning algorithms such as reinforcement
learning to predict future states of the networks, as well as suggest
possible optimal control actions.

\subsection{3D Reconstruction}\label{d-reconstruction}

Various 3D scanning technologies are emerging for a range of
applications, from outdoor surveying, 3D mapping of cities for digital
twins, inspection to autonomous driving. Most of these applications and
technologies rely on LiDAR sensors
\cite{liu2008airborne,nys2020automatic,sampath2009segmentation,wu2017graph}.
However, most LiDAR sensors tend to be expensive and heavy, making them
less suitable for developing a drone-based surveying solution. Other 3D
scanning solutions use a single lens
\cite{spreitzer2020large,santagati2013image} or stereo vision cameras
\cite{minos2018towards,henry2014rgb,choi2015robust,dryanovski2017large}
to compute a 3D model of the environment.

\subsubsection{Photogrammetry}\label{photogrammetry}

The most common method for 3D reconstruction of outdoor structures is
photogrammetry. The 3D representation of complex structures such as
buildings, bridges, and even 3D maps of a whole neighbourhood can be
generated using a single-lens camera based on the concept of Structure
from Motion (SfM) \cite{schonberger2016structure}.

The steps to create a point cloud or textured mesh is to capture
multiple photographs in sequence or randomising order with at least 70\%
overlapping and at angle part of around 5-10 degrees
\cite{santagati2013image}. This will ensure that the amount of overlap
is sufficient for matching photos to have common feature points.
Matching the features in different photos allows the SfM algorithm to
generate a 3D point cloud \cite{spreitzer2020large}. The generated point
cloud can be meshed to create a smooth or textured result of the 3D
model.

\subsubsection{Stereovision}\label{stereovision}

Stereovision is a 3D scanning method suitable for smaller or indoor
infrastructure projects where higher accuracy is required. The concept
uses stereovision cameras (infrared or RGB) to estimate the depth in the
field of view of the camera. Stereo Vision uses the disparity between
images from multiple cameras to extract depth information
\cite{Nair2012}. Similar to the binocular vision in humans, when both
eyes focus on an object, their optical axes will converge to that point
at an angle. The displacement parallel to the eye base (the distance
between both eyes) creates a disparity between both images. From the
extent of disparity, it is possible to extract the distance of an object
and pixel in an image through triangulation \cite{jgfryer2010}.

To generate the 3D model from the stereo or depth images, RGB-D cameras
require an additional processor to run a process called Simultaneous
Localisation and Mapping (SLAM) \cite{bevno3d}. As the name suggests,
the SLAM concept is able to build a 3D map of the environment in
real-time and at the same time estimate the location and orientation of
the camera. SLAM works by scanning the images for key features which can
be extracted with Speeded Up Robust Features (SURF)
\cite{bay2008speeded} and matched with RAndom SAmple Consensus (RASAC)
algorithm between multiple images \cite{fischler1981random}. These two
algorithms work simultaneously, SURF compares two images and extracts
matching key points. These key points are then combined with the depth
data to allow RANSAC algorithm to determine the 3D transformations
between the frames. The transformed key points are optimised into a
graph representation resulting in a 3D representation of the
environment.

\subsubsection{LiDAR Scanning}\label{lidar-scanning}

Laser measurements provide another means to obtain depth information of
the environment using the concept of time of flight of a light signal
reflected at the surrounding. Hence, LiDAR also uses an active approach
to obtain depth information similar to RGB-D technology. Still, LiDAR
sensors have a much larger range of 100 meters with accuracy in the
millimetre range. In recent years, LiDAR sensors have received a lot of
attention, mostly due to their extensive use in autonomous driving
technologies. This resulted in many available LiDAR sensors, which are
affordable and light enough to be installed on drones for aerial
scanning of infrastructure projects.

Similar to RGB-D sensors, most LiDAR-based 3D scanning techniques also
use a SLAM approach to convert the instantaneous laser-point
measurements to 2D or 3D point-cloud representations. GMapping is a
common SLAM technique introduced for LiDAR-based mapping, reducing the
computation time for the SLAM algorithms \cite{grisetti2007improved}.
HectorSLAM is the SLAM algorithm used here for the in-house development
of a 2D mapping evaluation \cite{kohlbrecher2011flexible}. It was first
developed for Urban Search and Rescue (USAR) scenarios and is suitable
for fast learning of occupancy grid maps with low computational
requirements. HectorSLAM presents a high update rate simultaneously on a
2D map for lower power platforms and the results yielded were a
sufficiently accurate mapping. A more recent SLAM algorithm by Google is
called Cartographer \cite{hess2016real}. In a comparison study
\cite{filipenko2018comparison}, GMapping produced an inaccurate mapping
while both the HectorSLAM and Cartographer produced accurate and similar
maps.

Many LiDAR-based 3D SLAM frameworks have been proposed specifically for
3D reconstruction and form the foundation for most commercial scanning
technologies available. Among the many LiDAR-based 3D SLAM methods, LOAM
is a widely used real-time LiDAR odometry estimation and mapping
framework that uses a LiDAR sensor and optionally an inertial
measurement unit (IMU) \cite{zhang2014loam}. This method achieves
real-time performance by separating the SLAM problem into odometry
estimation algorithm and mapping optimisation algorithm. The odometry
estimation algorithm runs at high frequency with low fidelity, while the
mapping optimization algorithm runs at an order of magnitude lower
frequency with high accuracy for scan-matching. Since its publication,
LOAM has remained at the top rank in the odometry category of various
benchmarks. LOAM has since then been commercialized, and its framework
is no longer available in the public domain.

The current state-of-the-art 3D SLAM method for LiDAR odometry and
mapping is LIO-SAM \cite{shan2020lio}. It utilizes factor graphs to
incorporate multiple measurement factors for odometry estimation and
global map optimization. The framework incorporates an IMU to improve
the pose estimation and incorporate GPS as an option for additional key
factors.

\section{Solution Design}\label{solution-design}

The backbone of our solution is an Information Fusion module to extend
beyond the current limitations of BIM. The Information Fusion module has
an extensive set of APIs, scalable storage, advanced search, and
indexing capabilities to fuse multiple data streams, capture different
types of BIM artefacts, AI models, and defects while supporting online
communication from our drones and management site.

For 3D reconstruction, we present our drone-based setup. The drone has a
stereovision camera attached as a cost-effective solution. The main
computing unit is a miniature onboard computer responsible for
processing the output from the camera feed via USB, and streaming it
back to the Information Fusion module in a real-time manner.

To test our defect detection use-case as an application of AI in
real-time image scanning, we deployed a deep learning model on the
on-board computer. The defects detected from the camera feeds are sent
back to the Information Fusion module to fuse with the 3D models and
other BIM-related information.

\begin{figure}
\centering
\includegraphics[width=1.0\textwidth]{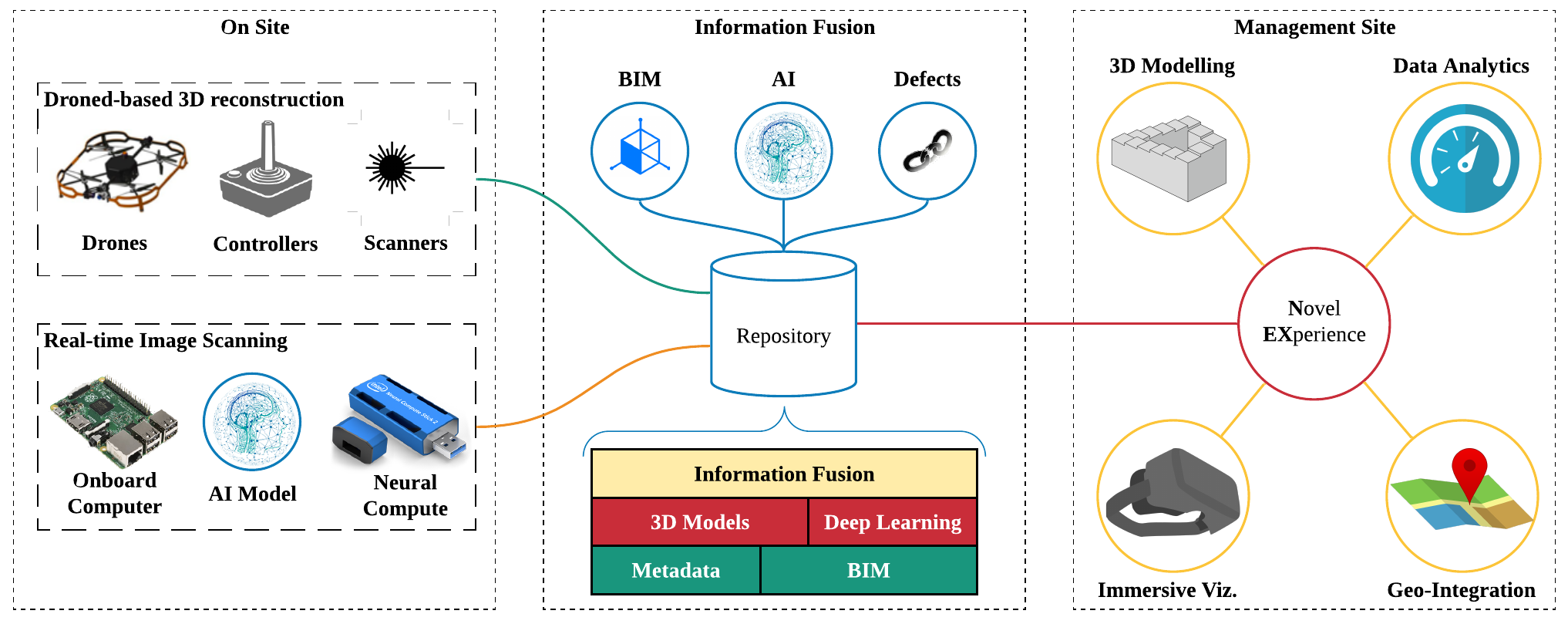}
\caption{Overall Architecture}
\label{fig:overall_architecture}
\end{figure}

The overall architecture is illustrated in
\figurename~\ref{fig:overall_architecture}. We also described each
component in detail in the following sections.

\subsection{Information Fusion}\label{information-fusion}

To be able to capture heterogeneous data sources including structured,
unstructured, images, 3D models, meta-information beyond BIM's
capabilities, The Information Fusion module leverages one of the most
efficient and well-established NoSQL database systems, Apache Cassandra,
originally developed by Facebook, hence, is able to handle a huge amount
of data across multiple locations, including on-site and off-site. With
an extended database schema, the module offers the ability to store and
replicate large BIM files with high data protection and fault-tolerance
while also supporting imaging data, defects, and tagged items. We also
added an extensive set of API to enable real-time communication from our
drone for live streaming of RGB-D images and defect information.

\subsection{Drone-based 3D
Reconstruction}\label{drone-based-3d-reconstruction}

\begin{figure}
\centering
\includegraphics[width=0.6\textwidth]{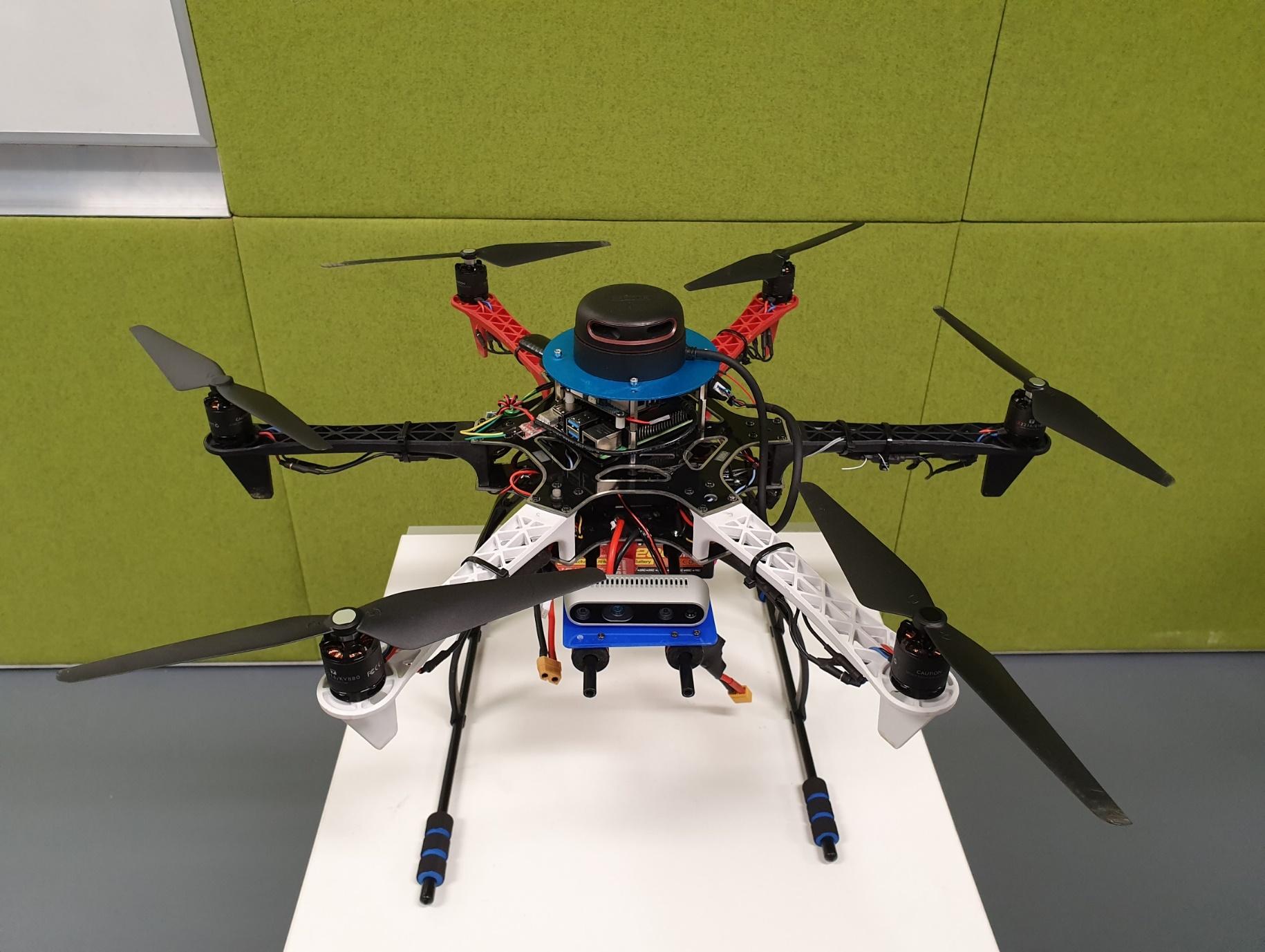}
\caption{Our drone configuration with integration of RPLidar A2 (bottom left) and Intel RealSense D435i (bottom right).}
\label{fig:drone}
\end{figure}

The stereovision camera used in our drone is an Intel RealSense D435i
camera which is more cost-effective compared to a LiDAR sensor. It is an
RGB-D camera that produces point-clouds in color instead of black and
white. The depth data provides the distance between the camera and the
obstacle in its FOV. It has an integrated Inertial Measurement Unit to
predict the orientation of the drone and provides a horizontal and
vertical FOV of 87 degrees by 58 degrees that allows a 3D map to be
generated. Our drone setup is shown in \figurename~\ref{fig:drone}

\subsubsection{Flight Controller}\label{flight-controller}

Each drone requires a flight controller to allow the pilot to have
precise control over the vehicle and its motors. Even in manual flight,
the flight controller translates the throttle command on the radio
control to individual motor commands to stabilize the drone. Flight
controllers use several inbuilt sensors to control the vehicle response.
In this work, we used a Pixhawk flight controller which allows us to
operate the custom drone. The Pixhawk flight controller also supports
many additional sensors and companion computers to be integrated.

\subsubsection{Onboard Computing}\label{onboard-computing}

We used a companion computer attached to our drone as the main
processing unit. In our prototype, a Raspberry Pi4 single-board computer
is added to allow additional sensors and features to be integrated. For
e.g.~it enables features such as obstacle avoidance, automated flight
path tracking, or in this work, 3D scanning of the environment.
Raspberry Pi4 is utilized in this prototype due to its low cost, high
specifications, and large supporting community. The other significant
factor for choosing the Raspberry Pi4 is its compatibility with the
additional sensors and the Robotic Operating Software (ROS) used. The
Intel RealSense D435i camera as mentioned in the previous sections is
connected and executed by the Raspberry Pi4 via USB port.

\subsubsection{Implementation of SLAM}\label{implementation-of-slam}

\begin{figure}
\centering
\includegraphics[width=0.8\textwidth]{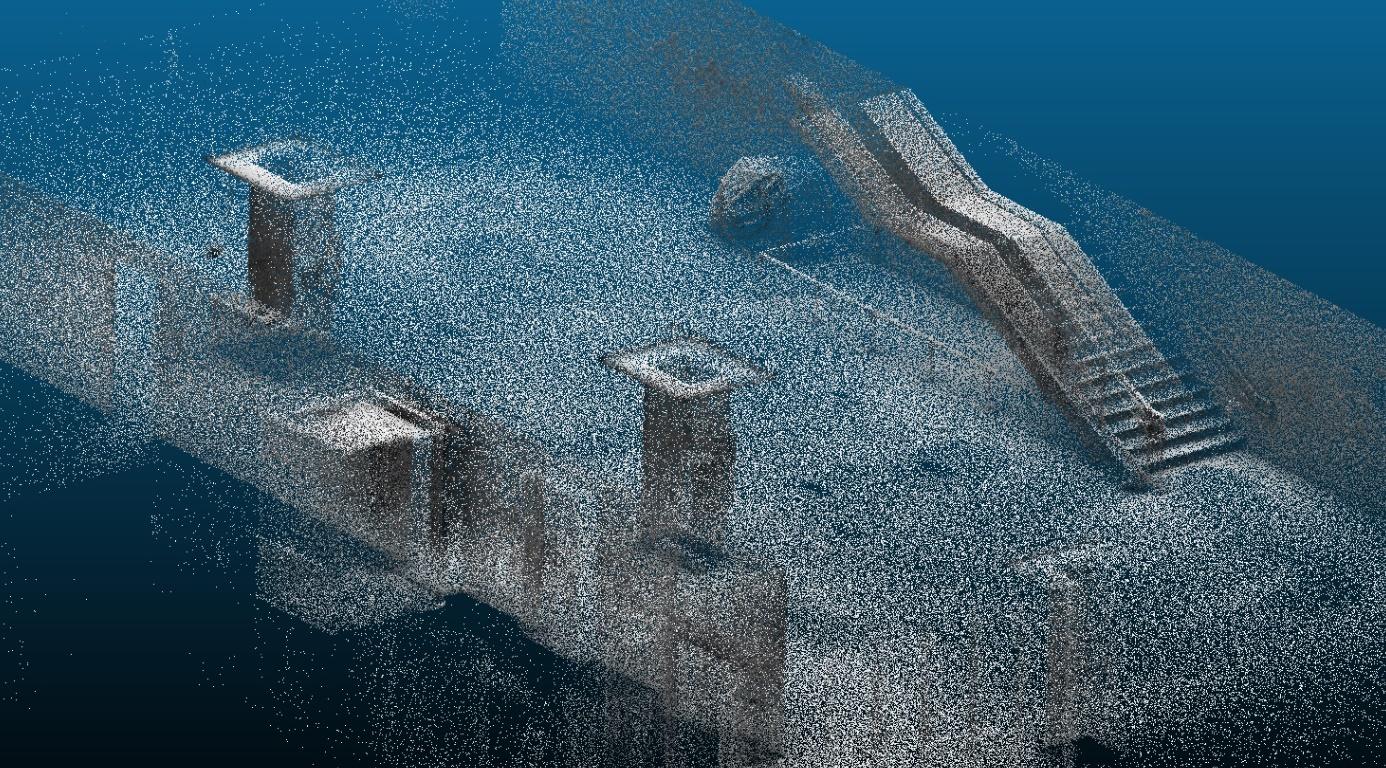}
\caption{3D reconstruction scanning.}
\label{fig:3drecon}
\end{figure}

Real-Time Appearance Based Mapping (RTABMap) is an open-source SLAM
environment \cite{labbe2019rtab} with numerous tools to generate maps
from RGB-D data. RTABMap has evolved to do online processing, minimal
drift odometry, robust localization map exploitation, and multi-session
mapping. The approach is based on the SLAM algorithm introduced before
and is illustrated in Figure \ref{fig:3drecon}, including the different
algorithms used to extract the features into the point cloud. Using SLAM
as the base to generate point clouds gives the user the flexibility to
change parameters or adjust flight paths during the scanning process.

\subsection{Real-time Image Scanning}\label{real-time-image-scanning}

One limitation of Raspberry Pi 4 as compared to conventional computers
is the limited processing power. This leads to low frame rates of only 5
frames per second. To provide more processing power to the Raspberry Pi
4 and allow more efficient Real-time Image Scanning, we explored the use
of USB accelerators to increase the frame rate. A USB accelerator is a
USB stick that contains a Vision Processing Unit aimed at boosting CPU
performance. The USB accelerator used in this work is an Intel Neural
Compute Stick 2 that is compatible with the Intel RealSense D435. It
also has a toolkit called the OpenVINO toolkit, which allows the
companion computer to recognise the NCS2 and make full use of the
additional CPU boost. After the implementation of the USB accelerator,
the frames rate provided a boost to the CPU of the Raspberry Pi 4
resulting in an average of 12 fps.

\subsection{AI for Defect Detection}\label{ai-for-defect-detection}

\begin{figure}[h!]
\centering
\includegraphics[width=.9\textwidth]{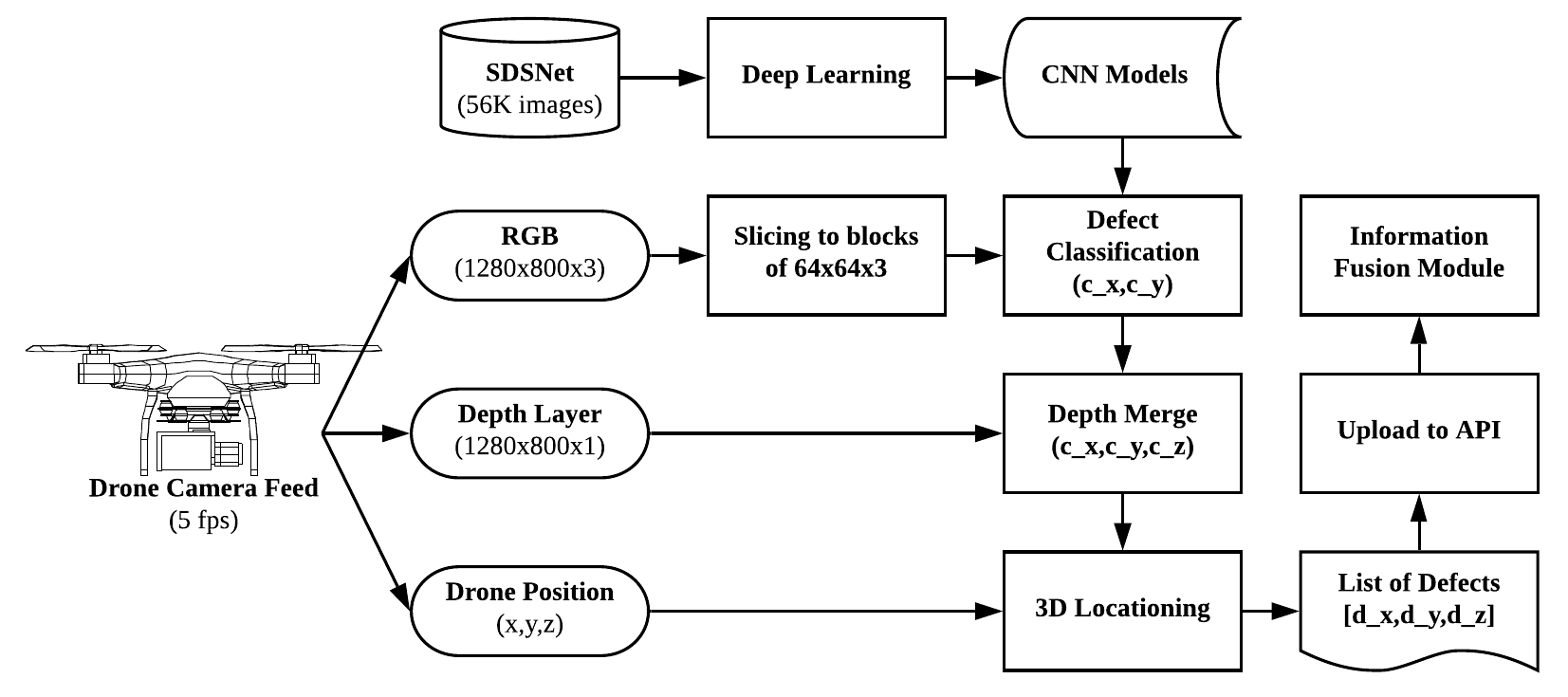}
\caption{AI Defect Detection}
\label{fig:crack_detection}
\end{figure}

To further evaluate our Real-time Image Scanning capability, we trained
a deep learning model for defect detection using convolutional neural
networks. We employed the SDNET 2018, a publicly available dataset, that
contains 56,000 images of cracks and non-cracks
\cite{dorafshan2018sdnet2018}. The dataset provides various types of
cracks, ranging from 0.06mm to 25mm, on different types of surfaces. We
trained our classifier engine with multiple backbones, including
ResNet18, ResNet50, and VGG; and then the classifiers' performance was
evaluated against current baselines. We utilized the best model to
classify 2D images, coming from streaming data sources. Our drone (in a
simulated environment) captures the Red-Green-Blue (RGB) channels and
the depth layer from RGBA images for processing. The drone position and
intrinsic camera can be configured to provide the best 3D locationing of
the defects for visualizing them in the simulation. The AI defect detection workflow is illustrated in \figurename\ \ref{fig:crack_detection}.

\section{Experiments and Results}\label{experiments-and-results}

We conducted experiments to validate our solution and answer the
following questions.

\begin{enumerate}
\item How do different scanning technologies perform compared to each other, and how do they perform compared to manual measurements?
\item How does a scanning technology perform when being used as a handheld device vs being used in a drone-based solution? And how are both approaches compared to manual measurements?
\item How do different CNN architectures perform in defect detection?
\end{enumerate}

The detail of the experiments and results are given in the following
sections.

\subsection{Scanning Performance}\label{scanning-performance}

\input{tab_3d_scanning_technologies}

We evaluated the performance of three different 3D scanning technologies
with the following specific products.

\begin{enumerate}
\item Photogrammetry with Pix4D
\item Stereovision with Dot3D/Navisworks
\item 3D LiDAR with geoSLAM/Navisworks.
\end{enumerate}

We manually measured selected areas of interest as well as scan them
with the listed products. For photogrammetry, we only included the
results for the ramp as the technology is deemed unsuitable for indoor
scanning. The results are summarised in Table
\ref{tab:3d_scanning_technologies}.

The evaluation of the methods showed that stereovision and 3D LiDAR
achieve accuracies sufficient for indoor surveying, with stereovision
achieving more consistent accuracies. Photogrammetry was found to not be suitable for indoor surveying
due to the high inaccuracy of the results.

\subsection{Measurement errors}\label{measurement-errors}

\subsubsection{Drone-based Inspection with
Stereovision}\label{drone-based-inspection-with-stereovision}

We used the drone-based setup described in Section \ref{solution-design} to
compare with manual measurements as well as when being used as a
handheld device. The results are given in Table
\ref{tab:stereovision_drone}.

\input{tab_stereovision_drone}

Both approaches produce very accurate results with the highest error of
1.3\%. In addition, the flight scan results are slightly improved even
since the drone only can move around in straight directions (up down,
left right, front back) for the scan to be completed. This means that
with lesser pitching of the drone the accuracy of results will be
improved. This demonstrates that the drone-based concept using
stereovision is a feasible approach for automated indoor scanning.

\subsubsection{Drone-based Inspection with 2D
LiDAR}\label{drone-based-inspection-with-2d-lidar}

Next, we compared handheld and drone-based scanning using the 2D LiDAR
approach against manual measurements. The result is given in Table
\ref{tab:lidar_drone}.

\input{tab_lidar_drone}

Similar to the stereovision approach, the drone-based scan for the 2D
LiDAR also shows better accuracy compared to the handheld scanning.
Although the difference between the handheld and drone-based readings is
small, with the largest being at around 1\%, it can be seen that the
drone-based scan produces more consistent results, as the drone is more
stable than the handheld method.

Comparison of the generated point-clouds from both scanning
technologies, stereovision, and LiDAR, shows that using a drone to
automate the scanning process has no detrimental effects. In fact, the
results demonstrate that drone-based scanning provides a more accurate
method compared to the handheld approach due to drone stability during
flight. Hence, our work demonstrated that it is possible to use 3D
scanning technologies integrated on a drone to enable automated indoor
surveying.

\subsection{Defect Detection
Performance}\label{defect-detection-performance}

\begin{figure}
\centering
\includegraphics[width=1.0\textwidth]{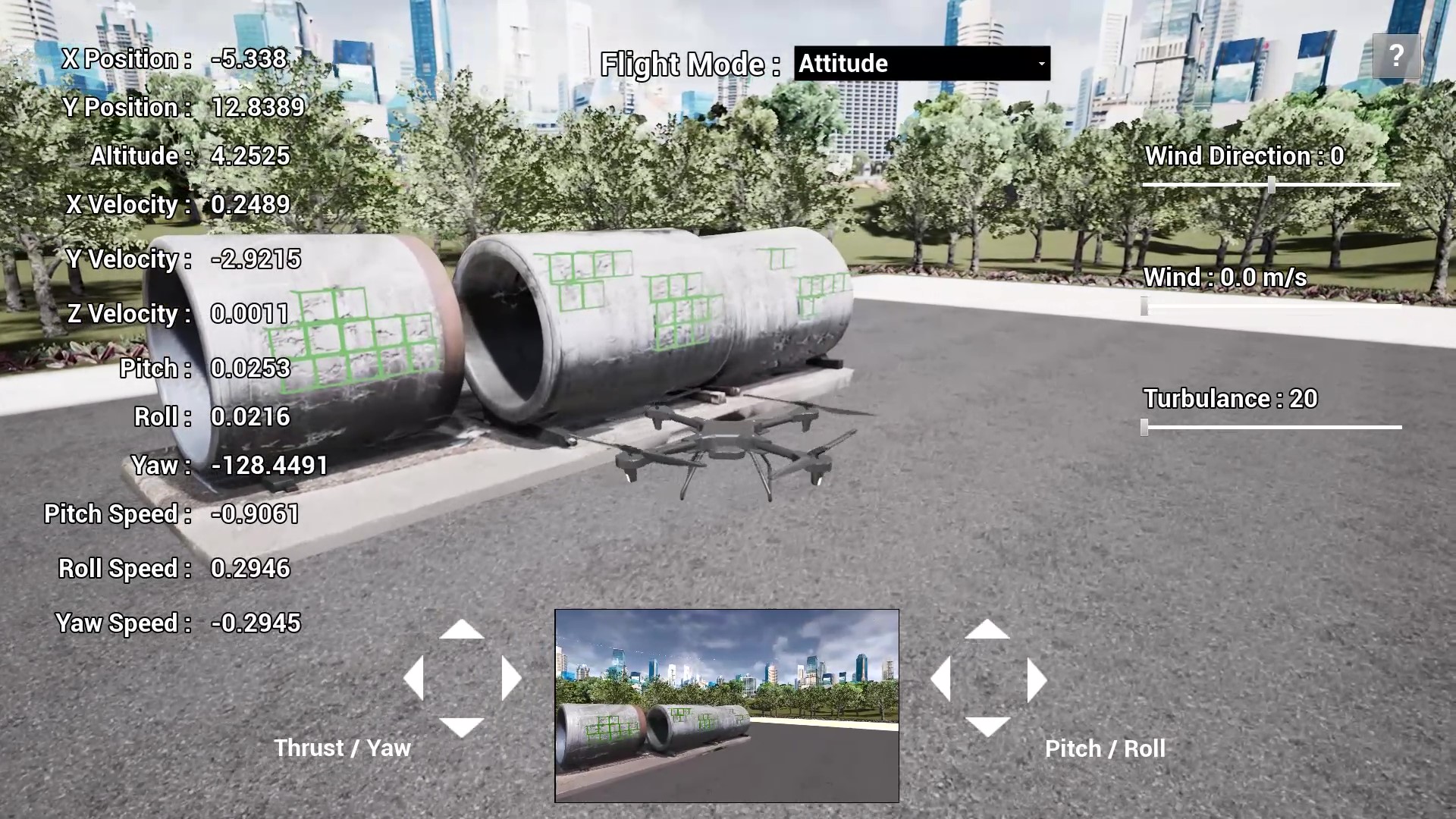}
\caption{Drone-based defect detection}
\label{fig:drone_crack_detection}
\end{figure}

Our drone-based setup scans the surrounding environment and uses the AI model deployed on the on-board computer for inference on the image stream as illustrated in \figurename\ \ref{fig:drone_crack_detection}.

The detection performance from our three trained models is given in the
Table \ref{tab:defect_perf}.

\input{tab_defect_perf}

The ResNet-50 outperformed ResNet-18 by 2\% in accuracy as well as a clear improvement of 9\% in the recall of crack detection.
The results showed that the deeper architecture allowed a better way to
recognise cracks in different forms. VGG-16 has achieved comparable
performance with ResNet-50. However, it has a slightly lower performance
in terms of F1-score in the Crack category, hence, Resnet-50 is selected
as our AI model of choice for defect detection

\section{Conclusion}\label{conclusion}

In this paper, we presented a drone-based AI and 3D Reconstruction for
Digital Twin augmentation. We illustrated an Information Fusion
framework that extends beyond BIM's capabilities to enable the
integration of heterogeneous data sources. We developed a proof of
concept drone-based 3D reconstruction and real-time image scanning and
provided evaluation and comparison results from extensive experiments.
Finally, we studied the feasibility of AI applications in real-time
image scanning through a defect detection use-case. Our work shows that
with Information Fusion, the applicability of BIM can be greatly
enhanced because the additional data allows additional applications such
as 3D reconstruction to be built on top of BIM. Our empirical
experiments also give suggestions to researchers and practitioners that
the use of drones, onboard computing, RGB-D cameras, and neural
computing unit are viable options for the realisation of large-scale,
real-time image scanning and AI in Digital Twin.

\section{Acknowledgement}\label{acknowledgement}

This research is conducted in collaboration with Aviation Virtual and
Nippon Koei.

%% file: tab_3D_scanning_technologies.tex
\begin{table}[b!]
\centering
\caption{3D Scanning Technologies}
\begin{tabular}{lrrrrrrr}
\toprule
                       & Actual                   & \multicolumn{2}{l}{Stereovision}                & \multicolumn{2}{l}{3D LiDAR}                    & \multicolumn{2}{l}{Photogrammetry}              \\
Area of Interest       & Dist.                    & M.                       & Rel. \%Err          & M.                       & Rel. \%Err          & M.                       & Rel. \%Err          \\
                       & (mm)                     & (mm)                     & \multicolumn{1}{l}{} & (mm)                     & \multicolumn{1}{l}{} & (mm)                     & \multicolumn{1}{l}{} \\
\midrule
\textbf{1) Ceiling height}      &                          &                          &                      &                          &                      &                          &                      \\
Ceiling height (L1)    & 3185                     & 3201                     & 0.5\%                & 3189                     & 0.1\%                &                          &                      \\
\midrule
\multicolumn{8}{l}{\textbf{2) Height of safety barriers}}                                                                                                                                                        \\
Balustrade (L2)        & 1120                     & 1111                     & -0.8\%               & 1007                     & -10.1\%              &                          &                      \\
Staircase Railing (L2) & 1110                     & 1082                     & -2.6\%               & 1018                     & -8.3\%               &                          &                      \\
\midrule
\multicolumn{8}{l}{\textbf{3) Profile of stairs}}                                                                                                                                                                \\
Thread width           & 1817                     & 1817                     & 0.0\%                & 1845                     & 1.5\%                &                          &                      \\
Riser height           & 148                      & 149                      & 0.7\%                & 157                      & 6.1\%                &                          &                      \\
\midrule
\multicolumn{8}{l}{\textbf{4) Dimensions of windows and doors}}                                                                                                                                                  \\
Lift door width        & 1195                     & 1199                     & 0.3\%                & 1204                     & 0.8\%                &                          &                      \\
Toilet door width      & 1135                     & 1139                     & 0.4\%                & 1116                     & 011.1\%              &                          &                      \\
Corridor width (L2)    & 2100                     & 2102                     & 0.1\%                & 1964                     & -6.5\%               &                          &                      \\
\midrule
\multicolumn{8}{l}{\textbf{5) Gradients of ramp}}                                                                                                                                                                \\
Ramp length            & 3800                     & 3845                     & 1.2\%                & 3851                     & 1.3\%                & 3890                     & 2.4\%                \\
Ramp height            & 295                      & 294                      & -0.3\%               & 297                      & 0.7\%                & 330                      & 11.9\%               \\
Ramp height/length     & 59/760                   & 64/837                   & -1.5\%               & 30/389                   & -0.7\%               & 33/389                   & 9.3\%               \\
\bottomrule
\end{tabular}
\label{tab:3d_scanning_technologies}

\end{table}

%% file: tab_stereovision_drone.tex
\begin{table}[]
\caption{Comparison between handheld and drone-based scanning using the stereovision approach against manual measurements}
\centering
\begin{tabular}{lrrrrrrr}
\toprule
                 & Measured & Handheld &        &          & Drone-based &        &          \\
Area of Interest & Distance & Distance & Error  & \% Error & Distance    & Error  & \% Error \\
\midrule
Room Width       & 9140 mm  & 9263 mm  & 123 mm & 1.34\%   & 9260 mm     & 120 mm & 1.31\%   \\
Shelf Width      & 690  mm  & 688 mm   & 2 mm   & 0.29\%   & 691 mm      & 1 mm   & 0.14\%   \\
Shelf Height     & 2130 mm  & 2127 mm  & 3 mm   & 0.14\%   & 2127 mm     & 3 mm   & 0.14\%  \\
\bottomrule
\end{tabular}
\label{tab:stereovision_drone}
\end{table}

%% file: tab_lidar_drone.tex
\begin{table}[]
\caption{Comparison between handheld and drone-based scanning using the 2D LiDAR approach against manual measurements}
\centering
\begin{tabular}{lrrrrrrr}
\toprule
                 & Measured      & Handheld &            &          & Drone-based &            &          \\
Area of Interest & Distance & Distance & Error & \% Error & Distance    & Error & \% Error \\
\toprule
Room Width       & 9140 mm          & 9103 mm          & 37 mm        & 0.40\%   & 9104 mm            & 36 mm        & 0.39\%   \\
Shelf Width      & 690 mm           & 695 mm           & 5 mm         & 0.72\%   & 697 mm             & 7 mm         & 1.01\%   \\
Door Height      & 950 mm           & 945 mm           & 5 mm         & 0.53\%   & 949 mm             & 1 mm         & 0.11\%  \\
\bottomrule
\end{tabular}
\label{tab:lidar_drone}
\end{table}

%% file: tab_defect_perf.tex
\begin{table}[]
\caption{Performance Evaluation of CNNs on Crack Detection}
\centering
\begin{tabular}{lrrrrrrr}
\toprule
         & Crack     &        &          & No Crack  &        &          & Overall Accuracy \\
         & Precision & Recall & F1-score & Precision & Recall & F1-score &                  \\
\midrule
Resnet18 & 0.86      & 0.52   & 0.65     & 0.92      & 0.98   & 0.95     & 0.91             \\
Resnet50 & 0.86      & 0.61   & 0.72     & 0.93      & 0.98   & 0.96     & 0.93             \\
VGG16    & 0.88      & 0.59   & 0.71     & 0.93      & 0.99   & 0.96     & 0.93             \\
\bottomrule
\end{tabular}
\label{tab:defect_perf}
\end{table}